# LLM SHOULD THINK AND ACTION AS A HUMAN

**Haun Leung** **ZiNan Wang**

## ABSTRACT

It is popular lately to train large language models to be used as chat assistants, but in the conversation between the user and the chat assistant, there are prompts, such as some commands or requests, require multi-turns between the chat assistant and the user. However, there are a number of issues with the multi-turns conversation: The response of the chat assistant is prone to errors and cannot help users achieve their goals, and as the number of conversation turns increases, the probability of errors will also increase; It is difficult for chat assistant to generate responses with different processes based on actual needs for the same command or request; Chat assistant require the use of tools to interactive with environment, but the current approach is not elegant and efficient enough, and due to the limited context window, the number of tool calls that can be supported is limited. The main reason for these issues is that large language models do not have the thinking ability as a human, lack the reasoning ability and planning ability, and lack the ability to execute plans. To solve these issues, we propose a thinking method based on a built-in chain of thought: In the multi-turns conversation, for each user prompt, the large language model thinks based on elements such as chat history, thinking context, action calls, memory and knowledge, makes detailed reasoning and planning, and actions according to the plan. We also explored how the large language model enhances thinking ability through this thinking method: Collect training datasets according to the thinking method and fine tune the large language model through supervised learning; Train a consistency reward model and use it as a reward function to fine tune the large language model using reinforcement learning, and the reinforced large language model outputs according to this way of thinking. Our experimental results show that the reasoning ability and planning ability of the large language model are enhanced, and the issues in the multi-turns conversation are solved.

## 1 Introduction

Recently, people have trained large language models as chat assistants, such as ChatGPT and Llama-3-chat [Touvron et al., 2023, Dubey et al., 2024], but in the conversation between the user and the chat assistant, there are prompts, such as some commands or requests, require multi-turns between the chat assistant and the user. [1] However, there are a number of issues with the multi-turns conversation.

**(1) The response of the chat assistant is prone to errors and cannot help users achieve their goals, and as the number of conversation turns increases, the probability of errors will also increase.** For example, when it comes to the user command 'Help me buy a birthday cake', large language models find it difficult to reason and plan, especially when executing the plan. Due to the involvement of multi-turns conversations, incorrect answers are likely to occur in the subsequent steps. If we consider the exceptions in tool calls during conversations and the user prompts noise in subsequent conversations, the responses of the large language model are more prone to errors.

**(2) It is difficult for chat assistant to generate responses with different processes based on actual needs for the same command or request.** For example, for the user command "Help me buy a birthday cake", sometimes we would like the response process to be "First, ask the user about the cake flavor, size, and budget, then use the ordering tool to place an order based on these parameters, and finally inform the user of the ordering result". But sometimes we hope that the response process is to "first recommend a few cakes to the user, including details and pictures, and ask the

[1] We call these commands and requests as action tasks.

user which one they like, then use the ordering tool to place an order based on the user's choice, and finally inform the user of the ordering result". We see that there may be many application scenarios, and it is difficult for large language models to generate responses for different processes based on actual needs.

**(3) Chat assistant require the use of tools to interactive with environment, but the current approach is not elegant and efficient enough, and due to the limited context window, the number of tool calls that can be supported is limited.** For example, it is common practice to put tool calls definitions in the system context. However, due to the limited length of the context window, an infinite number of tool calls are theoretically not supported. In addition, when the length of the system context becomes longer, the model inference time will become longer.

The main reason for these issues is that large language models do not have the thinking ability as a human, lack the reasoning ability and planning ability, and lack the ability to execute plans. To solve these issues, we propose a thinking method based on a built-in chain of thought: In the multi-turns conversation, for each user prompt, the large language model thinks based on elements such as chat history, thinking context, action calls, memory and knowledge, makes detailed reasoning and planning, and actions according to the plan. The chain of thought generated by the large language model according to this thinking method is built into the response and is wrapped by special tokens, commonly known as the built-in COTs.

We also explored how the large language model enhances thinking ability through this thinking method: Collect training datasets according to the thinking method and fine tune the large language model through supervised learning; Train a consistency reward model and use it as a reward function to fine tune the large language model using reinforcement learning, and the reinforced large language model outputs according to this way of thinking.

Our main contributions are four-fold.

**(1) The thinking method based on built-in chain of thought.** To solve the issues that exist in the large language model when completing action tasks, we propose a thinking method based on a built-in chain of thought. This thinking method defines the five elements on which the model thinks and the thought process. This thinking method sets the course for us to collect training datasets and ultimately enhances the model's thinking ability.

**(2) Use consistency reward model as reward function.** In multi-turns conversations, the large language model outputs responses based on user prompts and action calls results. How to judge the quality of the output responses has become the key to the reinforcement training effect. Usually, human preference reward models are used for judgment, but their accuracy is not high. Since the action task is not a problem of accurate reasoning, We can't use a rule-based reward system [DeepSeek-AI et al., 2025] either. To resolve the issue, we innovatively introduced consistency reward model that makes the consistency judgment for the output responses of the large language model. We found that the consistency reward model greatly enhances the effect of reinforcement training.

**(3) Solves the pain point of using system context.** System context is often used to store background information, as well as tool calls. Its disadvantages include: the system context will occupy the length of the context window, which limits the length of the output response; The limited length of the context window limits the number of function or tool calls; Excessive system context can slow down inference speed; Sometimes the answers of the model do not depend any tools at all, but they have already been loaded into the system context. We innovate the use of local thinking context, which can load background information and tool calls on demand, without causing these issues, and support an infinite number of tool calls.

**(4) Action calls are more elegant and efficient than tool calls.** We use action calls as tool calls. Compared to tool calls, it is syntactically elegant and action-efficient.

## 2 Related work

**Tool calls:** Function calls or tool calls of OepnAI ChatGPT. Langchain's prompt engineering enables the large language models to use tool calls. In [Li et al., 2024] propose a novel approach FnCTOD for solving DST with LLMs through function calling , this method improves zero-shot DST, allowing adaptation to diverse domains without extensive data collection or model tuning. Our job is to innovate the use of action calls, a more elegant and efficient tool calls.

**AI Agent:** Langchain's prompt engineering enables the large language models to solve problems. Andrew Ng has proposed design patterns specifically for building AI agents - Reflection, Tool Use, Planning, and Multi-Agent Collaboration. The focus of these studies is on AI agents, which use prompt engineering to prompt models to solve problems. In these jobs, AI agents are the active party, while the large language models are the passive party. Our focus is on the model, which is the active party, not the agent. The agent is only a tool provider to us, and how to use the tool and which tools to use are determined by the model. In other words, the agent is optional to us.

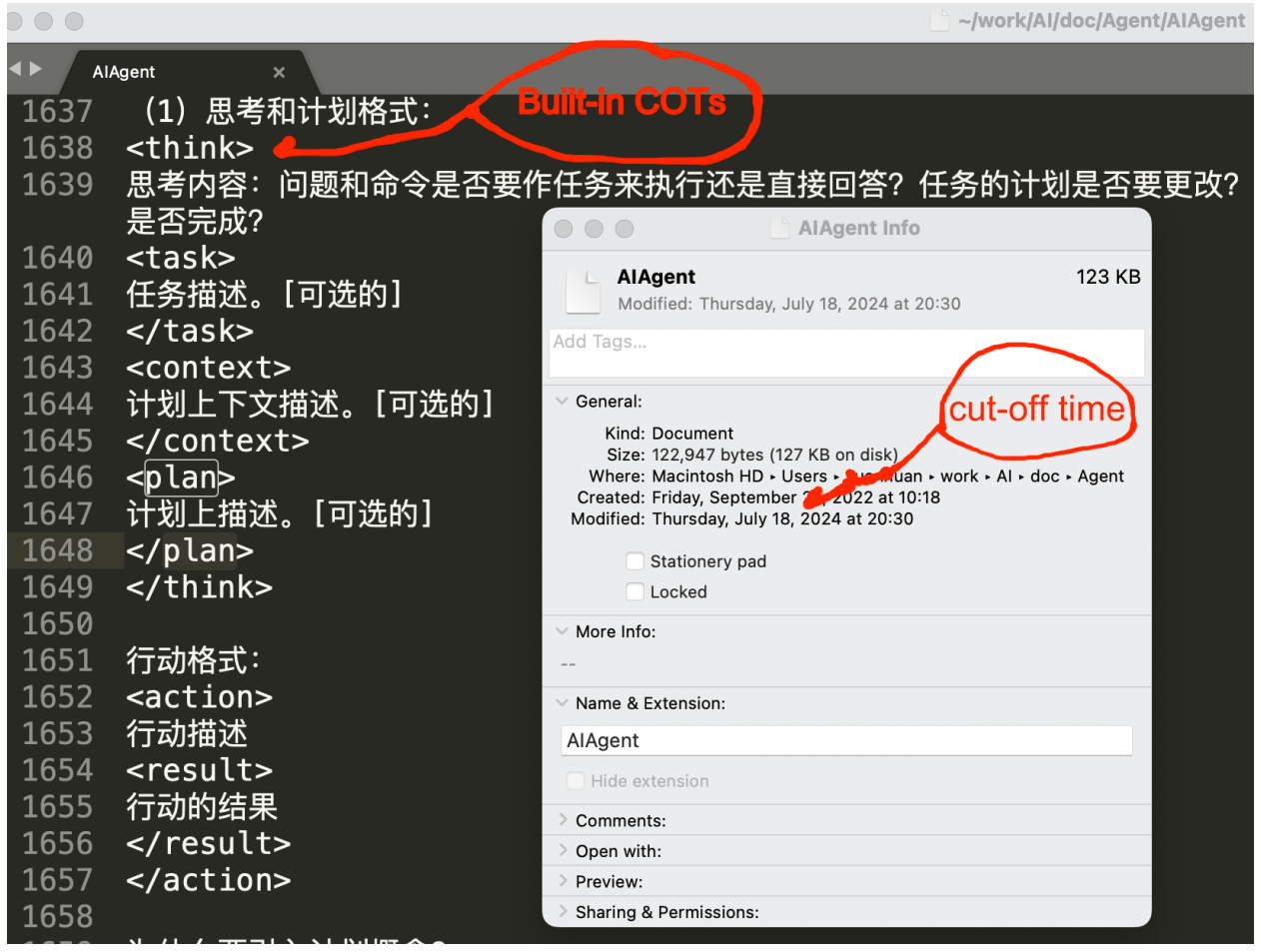

Figure 1: Time of using the built-in chain of thought.

**Reinforcement learning on large language models:** InstracutGPT [Ouyang et al., 2022] based reinforcement learning with human feedback and human preference reward model. The reinforcement learning pipeline of DeepSeeker R1 [DeepSeek-AI et al., 2025], as well as the "format and rule-based" reward function. We have benefited a lot from InstrcutGPT and DeepSeek-R1, the main difference between us and them is that the reward function is different, we innovate and use a consistent reward model.

**Built-in COTs:** The built-in chain of thought first appeared in the OpenAI o1-preview and o1-mini models, which were launched on September 12, 2024. Then on January 20, 2025, DeepSeek-R1 and DeepSeek-R1-Zero [DeepSeek-AI et al., 2025] also use built-in chain of thought. Our work also utilizes built-in chain of thought, but we need to emphasize that the draft of this paper was already in its infancy on July 18, 2024, when the built-in chain of thought was already in use, as stated in our paper title, "LLM Should think and action as a human". We have known since then that built-in chain of thought can enhance the reasoning and planning abilities of the large language models, while emphasizing that it must be encapsulated in special tokens. Evidence is shown in Figure 1.

**Conversations control over multi-turns:** Llama-2 [Touvron et al., 2023] proposes Ghost Attention (GAtt), a very simple method inspired by Context Distillation that hacks the fine-tuning data to help the attention focus in a multi-stage process. GAtt uses SFT training and is only for simple role-playing instructions. We work to reason and plan for complex contexts, and execute the plan, and we use RL training.

# 3 Method and experiment details

## 3.1 Overview

First, we propose a thinking method based on built-in chain of thought. Secondly, we collect an action tasks dataset based on this thinking method. Then, we use this action tasks dataset to fine tune the base model using supervise learning to obtain an initial strategy. Finally, we start with this initial strategy and conduct the reinforcement learning training process. The process then consists of three steps that can be repeated iteratively.

**Step 1: Collect samples from policies.** For each action tasks dataset sample, we sample responses from two sources including the current policy and initial policy. We then pair the action tasks dataset responses and sampling responses, and have a human evaluator evaluate the consistency to obtain consistency labels. We have obtained a consistency dataset.

**Step 2: Train a consistency reward model.** We use the consistency dataset to train a reward model to predict the logarithmic probability of consistency.

**Step 3: Optimize the policy according to the consistency reward model.** We treat the logit output of the consistency reward model as a reward that we optimize using reinforcement learning.

Finally, we explore conducting large-scale reinforcement learning on all tasks. We provide a more detailed description in the following section.

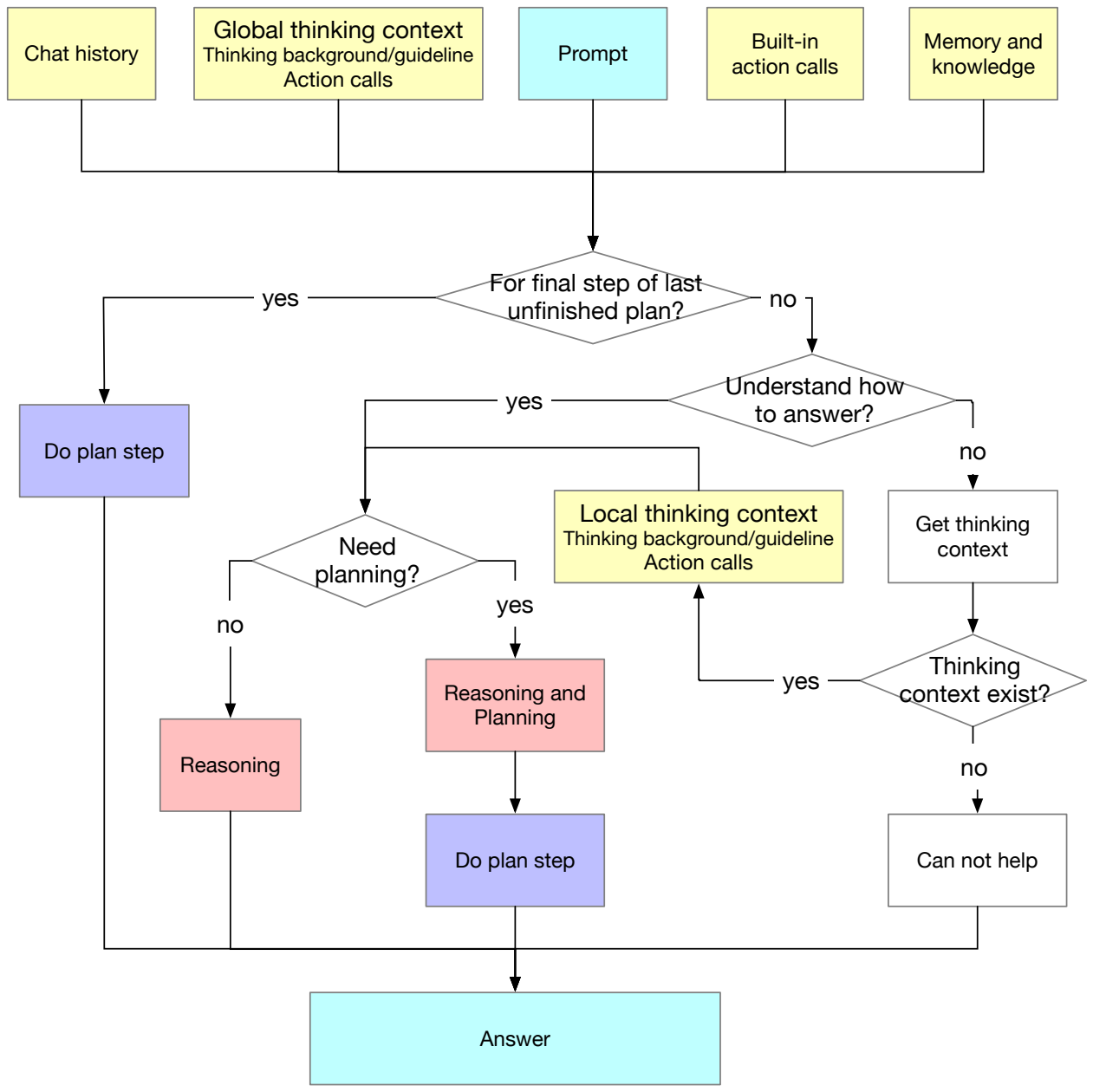

Figure 2: The thinking process of the thinking method.

### 3.2 Thinking method based on built-in chain of thought

#### 3.2.1 Thinking method

**Why innovate thinking method?** When handling with action tasks, large language models need to think carefully, makes detailed reasoning and planning, and actions according to the plan. But is the planning correct? Can all the planning steps be completed? How to interact with the environment? How to handle exceptions when using tool calls? During the execution of action tasks, How to do when a user initiates a new task? How to do when the user inputs noise that interferes with the correct process? If these complex questions are not handled well, the large language model will answer incorrectly. We needed a clear approach to what to do, so we propose a thinking method based on built-in chain of thought. The thinking method mainly consists of five elements, two processing logics and one planning step execution. The thinking process of the thinking method connects all these features together, shown in Figure 2.

**The thinking process of the thinking method.**
Based on the thinking elements: In multi-turns conversations, for each user prompt, the model first thinks based on four elements: chat history, global thinking context (Section 3.2.3), built-in action calls (Section 3.2.2), memory and knowledge.
The last plan step matches: If the user prompt is the last plan step corresponding to the last unfinished plan, then the model collects the useful information in the user prompt, and then proceeds to the current or next step of the plan, and if necessary, uses the action calls (Section 3.2.2) to interact with the environment, and deduces the answer based on the action calls result. Instead, the model thinks about how to answer based on the four elements.
Think about how to answer: If the model feels that it can not answer the prompt based on the four elements it already has, it further obtains the local thinking context (Section 3.2.3) related to the prompt. If the local thinking context does not exist, then the final answer is "can not help" or something like that; If exist, the next step will be moved to determine whether planning is needed. If the model feels that it can answer the prompt, it will move on to the next step to determine whether planning is needed.
Is planning needed? If the answer does not require a plan, then reason about it; Instead, reason and plan.
Reasoning processing logic: The model will conduct rigorous reasoning based on five elements, and if necessary, use action calls to interacte with environment and infer answer based on the results of action calls.
Reasoning and planning processing logic: The model will conduct rigorous reasoning and planning based on five elements, make a plan that includes multiple steps and begin executing the first step of the plan, and if necessary, use action calls to interacte with environment and infer answer based on the results of action calls.

**Built-in chain of thought.** Besides the answer, the thinking process of the thinking method is encapsulated in special tokens "«think»" and "«/think»", which is often referred to as built-in chain of thought. This is similar to OpenAI's GPT-o1-preview and Deepseek-R1.

**The elements of thinking method and their priorities.** There are five thinking elements: Chat History, Global thinking context, Built-in action calls, Local Thinking Context, Memory and Knowledge. Chat history refers to the history of all user prompts and model responses prior to the current user prompt. Memory and knowledge refers to the memory and knowledge that the model has, which is usually reflected in the training data. The ability of reasoning and planning when thinking is influenced by memory and knowledge. Global thinking context, Local thinking context, and Built-in action calls are described in the following sections. The priorities of thinking elements are as follows:

Chat History > Global thinking context > Built-in action calls > Local Thinking Context > Memory and Knowledge

When thinking, the model prioritizes high priority elements over low priority elements. For example, if there is an action call in the global thinking context that has the same functionality as a built-in action call, we will prioritize using the action call in the global thinking context.

**Details to pay attention to when thinking.**
Action call exception: When using action calls, various exceptions may be generated and returned. It is necessary to handle these exceptions well and try to ensure that the plan continues.
User initiated task interruption: In the current task of model processing, the user may actively interrupt the plan that is currently being executed.
User prompt noise: In the current task of model processing, users may input prompts that interfere with the process, and the model needs to think and respond.
Task nesting: In the current task of model processing, users may input prompts to perform new tasks, such as action tasks or other tasks.

### 3.2.2 Action calls and Built-in Action calls

Action calls are used when the model interacts with the environment, similar in function or tool calls. But action calls are more elegant and efficient than function calls or tool calls.

**Action calls definition.** Action call definition uses JSON structure, with attributes including: name, description, parameters, exception.

- name: The name of the action call, which is used by the model when thinking.
- description: A detailed description of the action call, including its purpose and usage scenarios.
- parameters: The description of the parameters passed in when using action calls, including but not limited to units, enumeration values, optional, etc.
- exception: The description of possible exceptions that may occur when using action calls.

For example, the action call definition for real-time weather query is as follows:

{ "name":"Check the real-time weather forecast.", "description":"Use to check the real-time weather forecast.", "parameters":[ "location":"The location to be queried. This's optional.", "time":"The time to be queried. This's optional.", ], "exception":"If the location does not exist, an exception will be returned." }

**Action calls usage.** Models use action calls when thinking. Format for using action call:

«action» The name of Action calls
Parameter 1: Value 1
Parameter 2: Value 2
...
Parameter n: Value n
«/action»
«observation»
The result of Action calls
«/observation»

| | |
|---|---|
| System | You are a helpful and thoughtful assistant.<br>You can order a birthday cake by the action calls bellow. You should handle possible exceptions in ordering and try to reorder as much as possible.<br>#action calls#<br>{"name":"Order a birthday cake by ordering system.", "description":"Use to order a birthday cake by ordering system.", "parameters":["flavor":"The flavor of cake.", "size":"The size of cake. There are three sizes: Small, Medium, or Large.", "budget":"The budget to buy cake. The unit is dollar."]} |
| User | I would like to buy a birthday cake. |
| Model | <<think>> I should help the user buy a birthday cake. Based on the global thinking context, the following planning steps apply:<br>step 1. Determine the type of cake: Ask user for more detail. This includes choosing the flavor and size of the cake.<br>step 2. Set a budget for the cake: Ask user for a budget.<br>step 3. Ask the user if they agree to place an order.<br>step 4. Place an order if user agrees: Use the ordering system to place an order; this is done by making an action call.<br>Let's do the step 1 first: Ask user to choose the flavor and size of the cake. <</think>><br>**What flavor of birthday cake do you like?** |
| System | You are a helpful and thoughtful assistant. |
| User | I would like to buy a birthday cake. |
| Model | <<think>> I should help the user buy a birthday cake. But I am not sure the way to do it. Let's first figure out how to do it.<br><<action>> Get thinking context.<br>content: how to buy a birthday cake.<br><</action>><br><<observation>> You can order a birthday cake by the action calls bellow. You should handle possible exceptions in ordering and try to reorder as much as possible.<br>#action calls#<br>{"name":"Order a birthday cake by ordering system.", "description":"Use to order a birthday cake by ordering system.", "parameters":["flavor":"The flavor of cake.", "size":"The size of cake. There are three sizes: Small, Medium, or Large.", "budget":"The budget to buy cake. The unit is dollar."]} <</observation>><br>Based on the observation, the following planning steps apply:<br>step 1. Determine the type of cake: Ask user for more detail. This includes choosing the flavor and size of the cake.<br>step 2. Set a budget for the cake: Ask user for a budget.<br>step 3. Ask the user if they agree to place an order.<br>step 4. Place an order if user agrees: Use the ordering system to place an order; this is done by making an action call.<br>Let's do the step 1 first: Ask user to choose the flavor and size of the cake. <</think>><br>**What flavor of birthday cake do you like?** |

Left: Global thinking context example
Right: Local thinking context example

Figure 3: Global thinking context and Local thinking context example.

**Action calls execution.** When the model uses action calls, the sampling algorithm stops sampling when generating "«/action»" tokens and hands over the string between "«action»" and "«/action»" tokens to the agent. The agent parses the action calls name, the parameters name and parameter values, executes the function corresponding to the action calls name, and returns the result generated by the function. The sampling algorithm concatenates the return result after the "«observation»" and ends with the "«/observation»" tokens. Then the model continues to think.

**Built-in Action calls.** We have built-in several action calls to facilitate model thinking. These built-in action calls are:

- Get thinking context
- Search Internet
- Open file
- Send Email
- Calculate

#### 3.2.3 Global thinking context and Local thinking context

**Global thinking context.** Global thinking context consists of two parts: Thinking background and guideline; Action calls definition. The thinking background is the data and information that the model depends on when thinking. The thinking guideline is a set of guiding rules for model thinking, typically used to guide how to create a plan. The action calls definition defines a list of action calls that the model can use to interact with the environment during thinking.

Global thinking context is placed in the system context. The role of global thinking context is similar to system context, guiding the model on how to do it and what to use to do it. As is well known, chat assistant such as ChatGPT typically place function calls and tool calls in the system context. The disadvantages of system context include: it can occupy the context window and limit the output length; Meanwhile, excessively long system context can lead to slower inference speed; Sometimes some input prompts do not require the use of any tools to call. To address these pain points, we have innovated local thinking contexts.

**Local thinking context.** Local thinking context consists of two parts: Thinking background and guideline; Action calls definition. The explanation of the thinking background and guidelines, and action calls definitions is the same as the global thinking context.

For each user input prompt, the model first considers four elements: chat history, global thinking context, built-in action calls, memory and knowledge. If the model feels that these four elements cannot help answer the prompt, it will further obtain local thinking context related to the prompt. The model obtains local thinking context through a built-in action calls.

| | | |
|---|---|---|
| System | You are a helpful and thoughtful assistant. | x0 |
| User | I would like to buy a birthday cake. | x1 |
| Model | <<think>> I should help the user buy a birthday cake. But I am not sure the way to do it. Let's first figure out how to do it.<br><<action>> Get thinking context.<br>content: how to buy a birthday cake.<br><</action>> | y1' |
| | <<observation>> You can order a birthday cake by the action calls bellow. You should handle possible exceptions in ordering and try to reorder as much as possible.<br>#action calls#<br>{"name":"Order a birthday cake by ordering system.", "description":"Use to order a birthday cake by ordering system.", "parameters": ["flavor":"The flavor of cake.", "size":"The size of cake. There are three sizes: Small, Medium, or Large.", "budget":"The budget to buy cake. The unit is dollar."]} <</observation>> | x2 |
| | Based on the observation, the following planning steps apply:<br>step 1. Determine the type of cake: Ask user for more detail. This includes choosing the flavor and size of the cake.<br>step 2. Set a budget for the cake: Ask user for a budget.<br>step 3. Ask the user if they agree to place an order.<br>step 4. Place an order if user agrees: Use the ordering system to place an order; this is done by making an action call.<br>Let's do the step 1 first: Ask user to choose the flavor and size of the cake. <</think>><br>**What flavor of birthday cake do you like?** | y2' |
| User | I like Lemon flavor. | x3 |
| Model | <<think>> User has chosen Lemon flavor. Let's continue the step 1. Ask user to choose the size of cake. <</think>><br>**What size cake is it? Small, Medium, or Large?** | y3' |
| User | Medium size is enough. | x4 |
| Model | <<think>> User has chosen Lemon flavor and Medium size for birthday cake. Let's do the step 2: Ask user for a budget. <</think>><br>**What is your budget for it?** | y4' |
| User | No more than 50$. | x5 |
| Model | <<think>> We have collected the information for ordering. Let's do the step 3: Ask the user if they agree to place an order. <</think>><br>**Okay, the ordering information is: Lemon flavor, Medium size, 50$ budget. Do you want to place an order right now?** | y5' |
| User | Okay, please. | x6 |
| Model | <<think>> The user has agreed to place the order. We now use the ordering system to place an order.<br><<action>> Order a birthday cake by ordering system.<br>flavor: Lemon<br>size: Medium<br>budget: 50$<br><</action>> | y6' |
| | <<observation>> Ordering successful. It will be delivered in about 48 hours. <</observation>> | x7 |
| | We have order the birthday successfully. <</think>><br>**Ordering successful and It will be delivered in about 48 hours.** | y7' |

Figure 4: Example of Action tasks dataset.

Global thinking context and Local thinking context example is shown in the Figure 3.

In practical applications, we recommend using local thinking context as the first choice, and do not recommend using global thinking context. We support global thinking context only to be compatible with the role of traditional system context.

### 3.2.4 Security

If there are illegal, violent, or pornographic contents in the thinking context, the model should refuse to execute it when thinking. Also If there are such contents in the result of the action calls, the model should ignore them and not display to the user.

## 3.3 Action tasks dataset collection

We collected an action tasks dataset $D$ containing thousands of samples, based on the thinking method. The collection ways include human labeler and prompt engineering based on high-performance language models.

**Sample format:** $x_1, y'_1, x_2, y'_2, ...., x_n, y'_n$. $x_i$ is user prompt or the results of action calls, $y'_i$ is label response. Example see Figure 4. Each sample also includes reference responses, that is in addition to label responses, the two responses are consistent in content and logic (see Section 3.5.1).

**Sample Distribution:** A variety of distributions should be taken into account when constructing the sample, and it is not limited to the following situations.
Action call exception: When using action calls, various exceptions may be generated and returned. It is necessary to handle these exceptions well and try to ensure that the plan continues.
User initiated task interruption: In the current task of model processing, the user may actively interrupt the plan that is currently being executed.
User prompt noise: In the current task of model processing, users may input prompts that interfere with the process, and the model needs to think and respond.
Task nesting: In the current task of model processing, users may input prompts to perform new tasks, such as action tasks or other tasks.
Application scenarios: food delivery, shopping, McDonald style ordering, device control, customer service ans so on.

### 3.4 Supervised Learning fine-tuning

We use this action tasks dataset to fine tune the base model with supervised learning, and obtain an initial policy.

### 3.5 Reinforcement Learning fine-tuning

#### 3.5.1 Collect samples from policies

**Samples collection.** For each sample of action tasks dataset $D$, we sample responses from two sources including the current policy and initial policy. For sample $(x_1, y_1', x_2, y_2', ...., x_n, y_n') \sim D$, $xi$ is user prompt or the results of action calls, $y_i'$ is label response. We use $(x_1, x_2, ..., x_n)$ to sample the policy $\pi$ and get the output responses $(y_1, y_2, ...., y_n)$. That is:

$$y_1 \sim \pi(x_1), y_2 \sim \pi(x_1, y_1, x_2), ...., y_n \sim \pi(x_1, y_1, x_2, y_2, ..., x_n)$$

**Consistency judgment.** We then pair the label responses of action tasks dataset sample and the sampling responses of policy: $[(y_1', y_1), (y_2', y_2), ...., (y_n', y_n)]$. And have a human evaluator evaluate the consistency, that is whether the sentences pair is consistent, and obtain consistency label $t$ equal to 0 or 1. Finally We have obtained a consistency dataset $D'$: $[(y_1', y_1, t_1), (y_2', y_2, t_2), ...., (y_n', y_n, t_n)] \sim D'$.

**The consistency of sentences pair.** If the content and logic of the two sentences is basically as the same, we say that it is consistent. But it is important to note that the planning steps that result from reasoning and planning cannot be shuffled, and if the order is not the same, we say that it is inconsistent.
For the example below, the result is consistent:

"What color do you like? The available colors are red, white, and blue."
"Pick up your favorite color from the available red, white, and blue color."

For the example below, the result is inconsistent:

"step 1: Pick up your favorite color. step 2: Choose the fit size. step 3: Place an order."
"step 1: Pick up your favorite color. step 2: Place an order. step 3: Choose the fit size."

#### 3.5.2 Train a consistency reward model

From the obtained consistency dataset $D'$, we train a reward model to predict the logarithmic probability of consistency. For performance reasons, we train the consistency reward model based on a high-performance small language model, such as Llama-3-8B.

**Implementation steps.** We remove the de-embedding layer of the Transformer decoder [Vaswani et al., 2017] and add a prediction header with an output dimension of 2. The logarithmic probability of consistency is given by the prediction header output. For each sample of consistency dataset $D'$: $[(y_1', y_1, t_1), (y_2', y_2, t_2), ...., (y_n', y_n, t_n)] \sim D'$, We packed the $(yi', yi)$ sentences pair in the sample into one prompt $x_i$, then train the reward model to predict consistency. The loss of consistency reward model is:

$$\text{loss}(r_\theta) = E_{(x,t)\sim D'}[cross_entropy_error(r_\theta(x), t)] \quad (1)$$

Where $r_\theta(x)$ is the output of the consistency reward model with parameter $\theta$. $cross_entropy_error$ is the cross entropy loss function. $x$ is prompt constructed from sentences pair. $t$ is consistency label.

**Prompt used for consistency prediction example.** Suppose there are such sentences pair in a sample of the consistency dataset, and the prompt $x$ is constructed as follows:

```
Question: Please check the consistency of the content described in the two sentences. If they are consistent, please answer yes. Otherwise, answer no. Answer without analysis, simply give yes or no.

## sentence 1 ##
"step 1: Pick up your favorite color. step 2: Choose the fit size. step 3: Place an order."

## sentence 2 ##
"step 1: Pick up your favorite color. step 2: Place an order. step 3: Choose the fit size."

Answer:
```

#### 3.5.3 Optimize the policy according to the consistency reward model

We use the trained consistency reward model to train a policy, and reinforce the thinking ability of policy. We initialize our policy as a model fine tuned using supervised learning on the action task dataset. For each sample of action tasks dataset: $(x_1, y'_1, x_2, y'_2, ...., x_n, y'_n) \sim D$, $xi$ is user prompt or the results of action calls, $y'_i$ is label response. We use $(x_1, x_2, ..., x_n)$ to sample the policy $\pi_\theta$ and get the output responses $(y_1, y_2, ...., y_n)$ as follows:

$$y_1 \sim \pi_\theta(x_1), y_2 \sim \pi_\theta(x_1, y_1, x_2), ...., y_n \sim \pi_\theta(x_1, y_1, x_2, y_2, ..., x_n)$$

We then optimize the policy by treating the output of the reward function as a reward for the policy output responses.

**Reward function.** The reward function is mainly composed of two types of rewards: format reward and consistency reward.
format reward: Check whether the special tokens in the policy output responses $(y_1, y_2, ...., y_n)$ comply with the rules. These special tokens includes "«think»", "«/think»", "«action»" and "«/action»".
consistency reward: Given by the consistency reward model. We encapsulate $[(y_1, y'_1), (y_2, y'_2), ..., (y_n, y'_n)]$ as prompts $[x_1, x_2, ..., x_n]$, and put them in batches into the consistency reward model to predict consistency and get the result $[r_1, r_2, ..., r_n]$. The ultimate consistency reward is $\prod_{i=1}^{n} r_i$.

### 3.6 Reinforcement Learning for all Tasks

In addition to action tasks, we can further integrate reasoning tasks and other tasks into the thinking method process. We have only analyzed possible implementation steps here, and the actual work will be left to the future.

**Action tasks.** Further collect more action tasks samples, in addition to human labeler and prompt engineering generation, trained policy can also be used to generate samples, but format filtering and manual screening are required. The training process involves supervised fine-tuning and RL fine-tuning on the base model, following the same logic as described above. Reward function is composed of two types of rewards: format reward and consistency reward.

**Reasoning tasks.** Similar to the approach of Deepseek-R1, reinforcement training is conducted for reasoning tasks with accurate answers. How to integrate it into the thinking method process of this paper? Equivalent to only one-turn of conversation. Reward function is composed of two types of rewards: format reward and rule-based reward.

**Other tasks.** Other tasks besides action tasks and reasoning tasks with accurate answers, such as writing tasks, knowledge tasks, and so on. How to integrate it into the thinking method process of this paper? Equivalent to only one-turn or multi-turns of conversation. Reward function is composed of two types of rewards: format reward and human preference reward.

Tasks and Reward function see Table 1.

Table 1: Tasks and Reward function

| Tasks | Reward function |
|---|---|
| Action tasks | format reward + consistency reward |
| Reasoning tasks | format reward + rule-based reward |
| Other tasks | format reward + human preference reward |

# 4 Results

## 4.1 The thinking ability of our model has been enhanced.

We found that our model's reasoning and planning capabilities, as well as the ability to execute the plan, were enhanced, by comparing with the baseline model. We use the action tasks test dataset, give the baseline model the same prompts and global thinking context and action call results, and then judge the model's completion rate of the tasks through human judgment and consistency reward model. We found that the baseline model did not have a higher completion rate for tasks than our model.

## 4.2 Action calls are more efficient than tool calls.

Action calls are syntactically more elegant than tool calls. Tool calls need to add two new message types, while action calls don't need to add any one. Action calls is just a message built into the chain of thought [Wei et al., 2022]. Moreover, the names of the action calls use the sentences that exist in human life, which are widely distributed in the training data. Action calls are also more efficient than tool calls. We use tool calls on the baseline model and action calls on our model, and use action tasks test datasets to fine tune both models with supervised training. Note that these tool calls and action calls have the same function. We observed that action calls performed better with the same amount of training data.

## 4.3 The relationship between reward model performance and dataset size and parameter size.

It will cost time too much for the consistency reward model to determine the consistency if the model parameters are too large, which will lead to a longer RL training time. So it is necessary to study the relationship between reward model performance and parameter size, and we will use the smallest model. How large the data size of the model needs to achieve ideal performance is also one of our concerns.

# 5 Discussion

## 5.1 Limitations.

One limitation of our work is that it is difficult to collect thinking context and cover multiple scenarios, which can lead to low generalization ability of the model. Collecting output from policies and evaluating consistency through manual labelers is also time-consuming and laborious, resulting in a smaller dataset for consistency action tasks. We need to think about how to improve the reinforcement algorithm, conduct more effective policy sampling and reinforcement training.

## 5.2 Future works.

In future work, we will explore the use of thinking method for large-scale reinforcement learning on various tasks. Integrate action tasks, reasoning tasks, and other tasks into the thinking method process and conduct large-scale reinforcement learning. Simultaneously consider how to collect more meaningful thinking contexts and cover a wider range of applications.